\documentclass[11pt,letterpaper]{article}
\usepackage{emnlp2016}
\usepackage{times}
\usepackage{latexsym}
\usepackage[hyphens]{url}
\usepackage[utf8]{inputenc}
\usepackage[english]{babel}
\usepackage[export]{adjustbox}[2011/08/13]
\usepackage{graphicx}
\usepackage{amsmath}
\usepackage{bm}
\usepackage{color}
\usepackage{xcolor}
\usepackage{scalerel}

\emnlpfinalcopy

\def\emnlppaperid{19}

\newcommand*{\emojiexample}{\scalerel*{\includegraphics{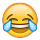}}}%

\newif\ifcomment
\commenttrue

\ifcomment
\newcommand{\mat}[1]{{\color{red}MB: #1}}
\newcommand{\ia}[1]{{\color{blue}IA: #1}}
\newcommand{\tim}[1]{{\color{green!75!black}TR: #1}}
\newcommand{\seb}[1]{{\color{purple}SR: #1}}

\else
\newcommand{\mat}[1]{}
\newcommand{\ia}[1]{}
\newcommand{\tim}[1]{}
\newcommand{\seb}[1]{}

\fi

\title{\verb~emoji2vec~: Learning Emoji Representations from their Description}

\author{
    Ben Eisner \\
    Princeton University \\
    {\tt beisner@princeton.edu}
  \And
    Tim Rocktäschel \\ 
    University College London \\ 
    {\tt t.rocktaschel@cs.ucl.ac.uk}
  \AND
    Isabelle Augenstein \\
    University College London \\
    {\tt i.augenstein@cs.ucl.ac.uk}
  \And
    Matko Bo\v{s}njak \\
    University College London \\
    {\tt m.bosnjak@cs.ucl.ac.uk}
  \AND
    Sebastian Riedel \\
    University College London \\
    {\tt s.riedel@cs.ucl.ac.uk}
}

\date{August 2016}

\begin{document}
\maketitle

\begin{abstract}
Many current natural language processing applications for social media
rely on representation learning and utilize pre-trained word embeddings.
There currently exist several publicly-available, pre-trained sets of word
embeddings, but they contain few or no emoji representations even as emoji
usage in social media has increased. In this paper we release
\verb~emoji2vec~, pre-trained embeddings for all Unicode emojis which are
learned from their description
in the Unicode emoji standard.\footnote{\url{http://www.unicode.org/emoji/charts/full-emoji-list.html}}
The resulting emoji embeddings can be readily used in downstream social natural language processing applications alongside \verb~word2vec~.
We demonstrate, for the downstream task of sentiment analysis, that
emoji embeddings learned from short descriptions outperforms a skip-gram
model trained on a large collection of tweets, while avoiding the need for
contexts in which emojis need to appear frequently in order to estimate a representation.
\end{abstract}

\section{Introduction}
First introduced in 1997, emojis, a standardized set of small pictorial glyphs depicting everything from smiling faces to international flags, have seen a drastic increase in usage in social media over the last decade.
The Oxford Dictionary named 2015 the year of the emoji, citing an increase in usage of over 800\% during the course of the year, and elected the `Face with Tears of Joy' emoji (\emojiexample))  as the Word of the Year.
As of this writing, over 10\% of Twitter posts and over 50\% of text on Instagram contain one or more emojis \cite{RefWorks:38}.\footnote{See \url{https://twitter.com/Kyle_MacLachlan/status/765390472604971009} for an extreme example.}
Due to their popularity and broad usage, they have been the subject of much formal and informal research in language and social communication, as well as in natural language processing (NLP).

In the context of social sciences, research has focused on emoji usage as a means of expressing emotions on mobile platforms. 
Interestingly, \newcite{kelly2015characterising} found that although essentially thought of as means of expressing emotions, emojis have been adopted as tools to express relationally useful roles in conversation. \cite{lebduska2014emoji} showed that emojis are culturally and contextually bound, and are open to reinterpretation and misinterpretation, a result confirmed by \cite{miller2016blissfully}.
These findings have paved the way for many formal analyses of semantic characteristics of emojis.

Concurrently we observe an increased interest in natural language processing on social media data~\cite{ritter-EtAl:2011:EMNLP,gattani2013entity,rosenthal-EtAl:2015:SemEval}. 
Many current NLP systems applied to social media rely on representation learning and word embeddings~\cite{tang-EtAl:2014:P14-1,dong-EtAl:2014:P14-2,dhingra-EtAl:2016:P16-2,augenstein2016stance}.
Such systems often rely on pre-trained word embeddings that can for instance be obtained from \verb~word2vec~~\cite{mikolov2013distributed} or \verb~GloVe~~\cite{pennington-socher-manning:2014:EMNLP2014}.
Yet, neither resource contain a complete set of Unicode emoji representations, which suggests that many social NLP applications could be improved by the addition of robust emoji representations.

In this paper we release \verb~emoji2vec~, embeddings for emoji Unicode symbols learned from their description in the Unicode emoji standard.
We demonstrate the usefulness of emoji representations trained in this way by evaluating on a Twitter sentiment analysis task.
Furthermore, we provide a qualitative analysis by investigating emoji analogy examples and visualizing the emoji embedding space.

\section{Related Work}
There has been little work in distributional embeddings of emojis. The first research done in this direction was an informal blog post by the Instagram Data Team in 2015 \cite{RefWorks:37}.
They generated vector embeddings for emojis similar to skip-gram-based vectors by training on the entire corpus of Instagram posts.
Their research gave valuable insight into the usage of emojis on Instagram, and showed that distributed representations can help understanding emoji semantics in everyday usage.
The second contribution, closest to ours, was introduced by \cite{emoji2016LREC}.
They trained emoji embeddings from a large Twitter dataset of over 100 million English tweets using the skip-gram method \cite{mikolov2013distributed}.
These pre-trained emoji representations led to increased accuracy on a similarity task, and a meaningful clustering of the emoji embedding space.
While this method is able to learn robust representations for frequently-used emojis, representations of less frequent emojis are estimated rather poorly or not available at all. In fact, only around 700 emojis can be found in \newcite{emoji2016LREC}'s corpus, while there is support of over 1600 emojis in the Unicode standard.

Our approach differs in two important aspects. First, since we are estimating the representation of emojis directly from their description, we obtain robust representations for all supported emoji symbols --- even the long tail of infrequently used ones. Secondly, our method works with much less data. Instead of training on millions of tweets, our representations are trained on only a few thousand descriptions. Still, we obtain higher accuracy results on a Twitter sentiment analysis task.

In addition, our work relates to the work of \newcite{hill2016learning} who built word representations for words and concepts based on their description in a dictionary. Similarly to their approach, we build representations for emojis based on their descriptions and keyword phrases.

Some of the limitations of our work are evident in the work of \newcite{park2013emoticon} who showed that different cultural phenomena and languages may co-opt conventional emoji sentiment. Since we train only on English-language definitions and ignore temporal definitions of emojis, our training method might not capture the full semantic characteristics of an emoji.

\section{Method}
\begin{figure}[t!]
\centering
\fbox{\includegraphics[width=0.95\columnwidth]{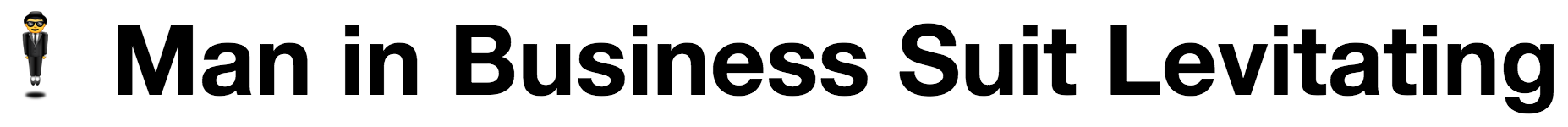}}
\caption{Example description of U+1F574. We also use \emph{business}, \emph{man} and \emph{suit} keywords for training.}
\label{fig:ex}
\end{figure}

Our method maps emoji symbols into the same space as the 300-dimensional Google News \verb~word2vec~ embeddings.
Thus, the resulting \verb~emoji2vec~ embeddings can be used in addition to 300-dimensional \verb~word2vec~ embeddings in any application.
To this end we crawl emojis, their name and their keyword phrases from the Unicode emoji list, resulting in 6088 descriptions of 1661 emoji symbols.
Figure \ref{fig:ex} shows an example for an uncommon emoji.

\subsection{Model}
We train emoji embeddings using a simple method. 
For every training example consisting of an emoji and a sequence of words $w_1, \ldots, w_N$ describing that emoji, we take the sum of the individual word vectors in the descriptive phrase as found in the Google News \verb~word2vec~ embeddings
$$ \bm{v}_j = \sum_{k=1}^N \bm{w}_{k} $$
where $\bm{w}_k$ is the \verb~word2vec~ vector for word $w_k$ if that vector exists (otherwise we drop the summand) and $\bm{v}_j$ is the vector representation of the description.
We  define a trainable vector $\bm{x}_i$ for every emoji in our training set, and model the probability of a match between the emoji representation $\bm{x}_i$ and its description representation $\bm{v}_j$ using the sigmoid of the dot product of the two representations $\sigma(\bm{x}_i^T\bm{v}_j)$.
For training we use the logistic loss
$$ \mathcal{L}(i,j,y_{ij}) = -\log(\sigma(y_{ij}\bm{x}_i^T\bm{v}_j-(1-y_{ij})\bm{x}_i^T\bm{v}_j))$$
where $y_{ij}$ is $1$ if description $j$ is valid for emoji $i$ and $0$ otherwise.

\subsection{Optimization}
Our model is implemented in TensorFlow \cite{abadi2015tensorflow} and optimized using stochastic gradient descent with Adam \cite{kingma2015adam} as optimizer.
As we do not observe any negative training examples (invalid descriptions of emojis do not appear in the original training set), to increase generalization performance we randomly sample descriptions for emojis as negative instances (i.e. induce a mismatched description).
One of the parameters of our model is the ratio of negative samples to positive samples; we found that having one positive example per negative example produced the best results.
We perform early-stopping on a held-out development set and found 80 epochs of training to give the best results. As we are only training on emoji descriptions and our method is simple and cheap, training takes less than 3 minutes on a 2013 MacBook Pro.

\section{Evaluation}
We quantitatively evaluate our approach on an intrinsic (emoji-description classification) and extrinsic (Twitter sentiment analysis) task. Furthermore, we give a qualitative analysis by visualizing the learned emoji embedding space and investigating emoji analogy examples.

\subsection{Emoji-Description Classification}
To analyze how well our method models the distribution of correct emoji descriptions, we created a manually-labeled test set containing pairs of emojis and phrases, as well as a correspondence label. For instance, our test set includes the example: \{\emojiexample), "crying", True\}, as well as the example \{\emojiexample), "fish", False\}. We calculate $\sigma(\bm{x}_i^T\bm{v}_i)$ for each example in the test set, measuring the similarity between the emoji vector and the sum of word vectors in the phrase.

When a classifier thresholds the above prediction at $0.5$ to determine a positive or negative correlation, we obtain an accuracy of 85.5\% for classifying whether an emoji-description pair is valid or not. 
By varying the threshold used for this classifier, we obtain a receiver operating characteristic curve (Figure~\ref{fig:ROC}) with an area-under-the-curve of 0.933, which demonstrates that high quality of the learned emoji representations.

\begin{figure}[t]
\centering
\includegraphics[width=1.0\columnwidth, center]{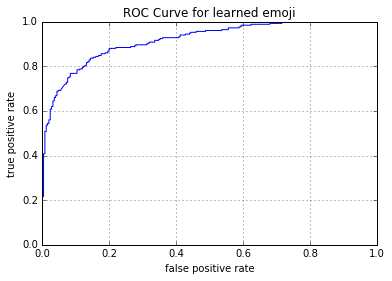}
\caption{Receiver operating characteristic curve for learned emoji vectors evaluated against the test set.
}
\end{figure}\label{fig:ROC}

\subsection{Sentiment Analysis on Tweets}
As downstream task we compare the accuracy of sentiment classification of tweets for various classifiers with three different sets of pre-trained word embeddings: (1) the original Google News \verb~word2vec~ embeddings, (2) \verb~word2vec~ augmented with emoji embeddings trained by~\newcite{emoji2016LREC}, and (3) \verb~word2vec~ augmented with \verb~emoji2vec~ trained from Unicode descriptions. 
We use the recent dataset by \newcite{POne:1}, which consists of over 67k English tweets labelled manually for positive, neutral, or negative sentiment. In both the training set and the test set, 46\% of tweets are labeled neutral, 29\% are labeled positive, and 25\% are labeled negative. To compute the feature vectors for training, we summed the vectors corresponding to each word or emoji in the text of the Tweet. 
The goal of this simple sentiment analysis model is not to produce state-of-the-art results in sentiment analysis; it is simply to show that including emojis adds discriminating information to a model, which could potentially be exploited in more advanced social NLP systems.

\begin{table*}[!ht]
\centering
\begin{tabular}{ |p{6cm}||p{3cm}|p{3.5cm}|  }
 \hline
 \multicolumn{3}{|c|}{Classification accuracy on entire dataset, \(N=12920\)} \\
 \hline
 Word Embeddings & Random Forest & Linear SVM \\
 \hline

 Google News & 57.5 & 58.5 \\
 Google News + \cite{emoji2016LREC} & 58.2\textsuperscript{*} & 60.0\textsuperscript{*} \\
 Google News + \verb~emoji2vec~ & \bf 59.5\textsuperscript{*} & \bf 60.5\textsuperscript{*} \\
 \hline
 \hline
 \multicolumn{3}{|c|}{Classification accuracy on tweets containing emoji, \(N=2295\)} \\
 \hline
 Word Embeddings & Random Forrest & Linear SVM \\
 \hline

 Google News & 46.0 & 47.1 \\
 Google News + \cite{emoji2016LREC} & 52.4\textsuperscript{*} & 57.4\textsuperscript{*} \\
 Google News + \verb~emoji2vec~ & \bf 54.4\textsuperscript{*} & \bf 59.2\textsuperscript{*} \\
 \hline
 \hline
 \multicolumn{3}{|c|}{Classification accuracy on 90\% most frequent emoji, \(N=2186\)} \\
 \hline
 Word Embeddings & Random Forrest & Linear SVM \\
 \hline
 Google News & 47.3 & 45.1 \\
 Google News + \cite{emoji2016LREC} & 52.8\textsuperscript{*} & 56.9\textsuperscript{*} \\
 Google News + \verb~emoji2vec~ & \bf 55.0\textsuperscript{*} & \bf 59.5\textsuperscript{*} \\
 \hline
 \hline
 \multicolumn{3}{|c|}{Classification accuracy on 10\% least frequent emoji, \(N=308\)} \\
 \hline
 Word Embeddings & Random Forrest & Linear SVM \\
 \hline
 Google News & 44.7 & 43.2 \\
 Google News + \cite{emoji2016LREC} & 53.9\textsuperscript{*} & 52.9\textsuperscript{*} \\
 Google News + \verb~emoji2vec~ & \bf 54.5\textsuperscript{*} & \bf 55.2\textsuperscript{*} \\
\hline
\end{tabular}
\caption{Three-way classification accuracy on the Twitter sentiment analysis corpus using Random Forrests \protect\cite{ho1995random} and Linear SVM \protect\cite{fan2008liblinear} classifier with different word embeddings. "*" denotes results with significance of $p<0.05$ as calculated by McNemar's test, with the respect to classification with Google News embeddings per each classifier, and dataset}
\label{table:1}
\end{table*}

Because the labels are rather evenly distributed, accuracy is an effective metric in determining performance on this classification task. Results are reported in Table \ref{table:1}. 
We find that augmenting \verb~word2vec~ with emoji embeddings improves overall classification accuracy on the full corpus, and substantially improves classification performance for tweets that contain emojis.
It suggests that emoji embeddings could improve performance for other social NLP tasks as well.
Furthermore, we find that \verb~emoji2vec~ generally outperforms the emoji embeddings trained by \newcite{emoji2016LREC}, despite being trained on much less data using a simple model.

\subsection{t-SNE Visualization}
To gain further insights, we project the learned emoji embeddings into two-dimensional space using t-SNE~\cite{maaten2008visualizing}. 
This method projects high-dimensional embeddings into a lower-dimensional space while attempting to preserve relative distances.
We perform this projection of emoji representation into two-dimensional space. 

From Figure~\ref{fig:t-sne} we see a number of notable semantic clusters, indicating that the vectors we trained have accurately captured some of the semantic properties of the emojis.
For instance, all flag symbols are clustered in the bottom, and many smiley faces in the center. Other prominent emoji clusters include fruits, astrological signs, animals, vehicles, or families. 
On the other hand, symbolic representations of numbers are not properly disentangled in the embedding space, indicating limitations of our simple model. A two-dimensional projection is convenient from a visualization perspective, and certainly shows that some intuitively similar emojis are close to each other in vector space.

\begin{figure*}[!ht]
\centering
\includegraphics[width=1.0\textwidth, center]{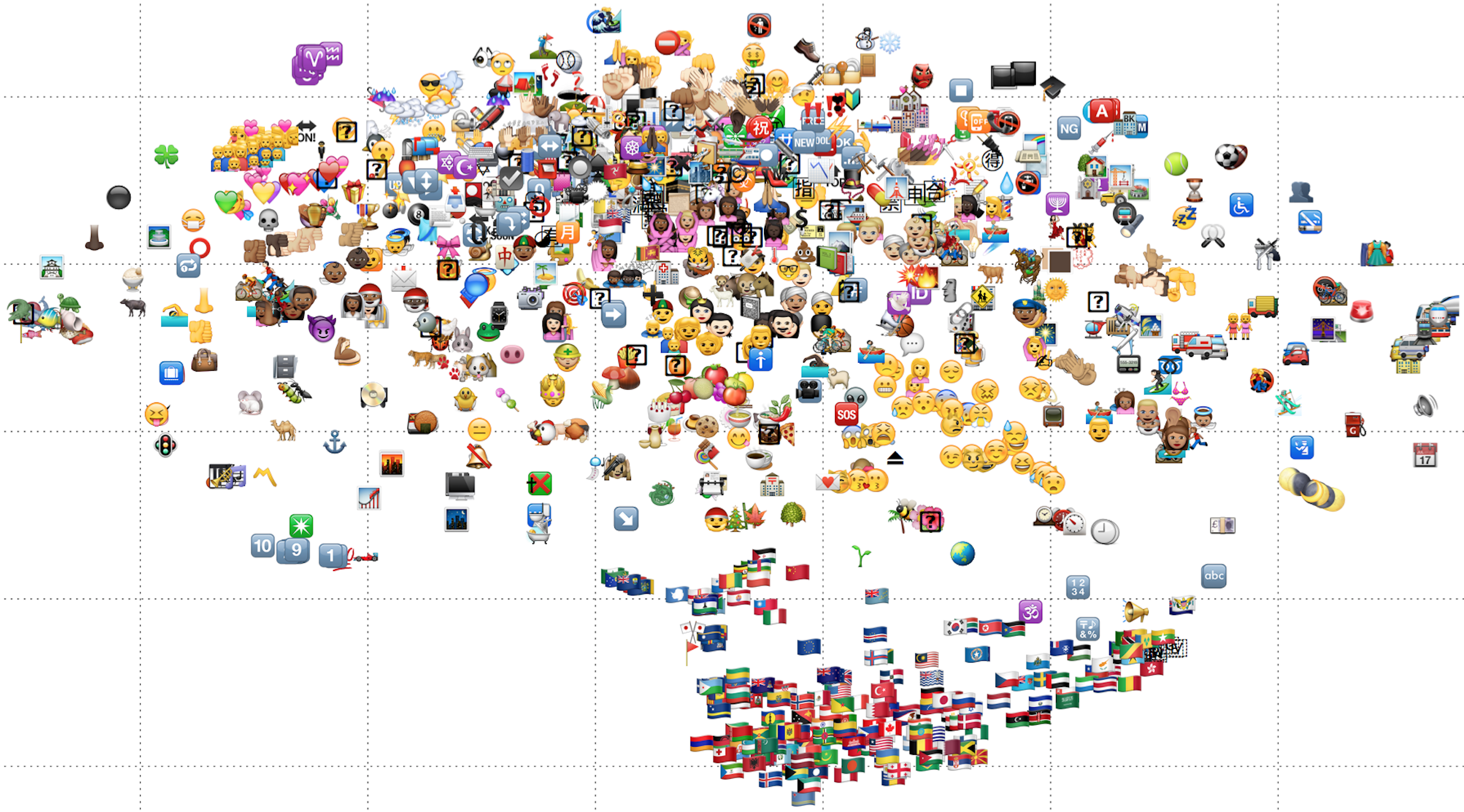}
\caption{Emoji vector embeddings, projected down into a 2-dimensional space using the t-SNE technique. Note the clusters of similar emojis like flags (bottom), family emoji (top left), zodiac symbols (top left), animals (left), smileys (middle), etc.}
\end{figure*}\label{fig:t-sne}

\subsection{Analogy Task}
\begin{figure}[!ht]
\centering
\includegraphics[width=1.0\columnwidth, center]{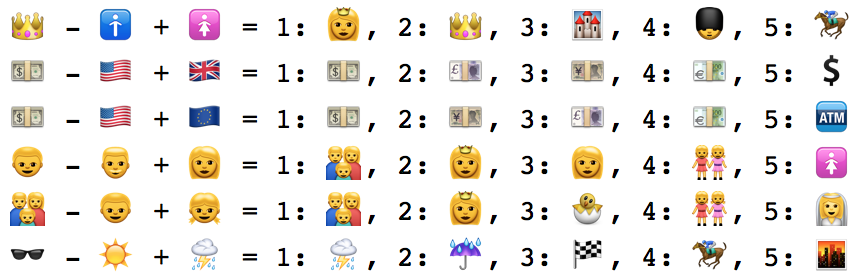}
\caption{Emoji analogy exmaples. Notice that the seemingly "correct" emoji often appears in the top three closest vectors, but not always in the top spot (furthest to the left).
}
\label{fig:analogy}
\end{figure}

A well-known property of \verb~word2vec~ is that embeddings trained with this method to some extent capture meaningful linear relationships between words directly in the vector space. 
For instance, it holds that the vector representation of 'king' minus 'man' plus 'woman' is closest to 'queen'~\cite{mikolov-yih-zweig:2013:NAACL-HLT}.
Word embeddings have commonly been evaluated on such word analogy tasks \cite{levy-goldberg:2014:W14-16}.
Unfortunately, it is difficult to build such an analogy task for emojis due to the small number and semantically distinct categories of emojis.
Nevertheless, we collected a few intuitive examples in Figure~\ref{fig:analogy}. 
For every query we have retrieved the closest five emojis.
Though the correct answer is sometimes not the top one, it is often contained in the top three.

\section{Conclusion}
Since existing pre-trained word embeddings such as Google News \verb~word2vec~ embeddings or \verb~GloVe~ fail to provide emoji embeddings, we have released \verb~emoji2vec~ --- embeddings of 1661 emoji symbols.
Instead of running \verb~word2vec~'s skip-gram model on a large collection of emojis and their contexts appearing in tweets, \verb~emoji2vec~ is directly trained on Unicode descriptions of emojis.
The resulting emoji embeddings can be used to augment any downstream task that currently uses \verb~word2vec~ embeddings, and might prove especially useful in social NLP tasks where emojis are used frequently (\emph{e.g.} Twitter, Instagram, etc.).
Despite the fact that our model is simpler and trained on much less data, we outperform \cite{emoji2016LREC} on the task of Twitter sentiment analysis.

As our approach directly works on Unicode descriptions, it is not restricted to emoji symbols. In the future we want to investigate the usefulness of our method for other Unicode symbol embeddings.
Furthermore, we plan to improve \verb~emoji2vec~ in the future by also reading full text emoji description from Emojipedia\footnote{\url{emojipedia.org}} and using a recurrent neural network instead of a bag-of-word-vectors approach for enocoding descriptions. 
In addition, since our approach does not capture the context-dependent definitions of emojis (such as sarcasm, or appropriation via other cultural phenomena), we would like to explore mechanisms of efficiently capturing these nuanced meanings.

\section*{Data Release and Reproducibility}
Pre-trained \verb~emoji2vec~ embeddings as well as the training data and code are released at \url{https://github.com/uclmr/emoji2vec}.
Note that the \verb~emoji2vec~ format is compatible with \verb~word2vec~ and can be loaded into gensim\footnote{\url{https://radimrehurek.com/gensim/models/word2vec.html}} or similar libraries.

\section*{Acknowledgements}
The authors would like to thank Michael Large, Peter Gabriel and Suran Goonatilake for the inspiration of this work, and the anonymous reviewers for their insightful comments.
This research was supported by an Allen Distinguished Investigator award, a Marie Curie Career Integration Award, by Microsoft Research through its PhD Scholarship Programme, and by Elsevier.

\bibliography{main}
\bibliographystyle{emnlp2016}

\end{document}